\documentclass[10pt,twocolumn,letterpaper]{article}

\usepackage{cvpr}              
\usepackage{placeins}
\usepackage{listings}
\usepackage{geometry}
\usepackage{xcolor}
\usepackage{amsmath}
\usepackage{tikz}
\usepackage{pgfplots}
\pgfplotsset{compat=1.18}
\usepgfplotslibrary{groupplots}

\definecolor{color_robot}{RGB}{248, 244, 225}
\definecolor{color_robot_human_no_skeleton}{RGB}{49, 49, 98}
\definecolor{color_robot_human_no_pretrain}{RGB}{158, 226, 248}
\definecolor{color_ours}{RGB}{124, 186, 165}

\usepackage[acronym]{glossaries}
\makeglossaries
\newacronym{vla}{VLA}{vision-language-action}
\newacronym{il}{IL}{imitation learning}
\newacronym{rl}{RL}{reinforcement learning}

\usepackage{pgfplots}
\pgfplotsset{compat=1.18}
\usepackage{amsmath}   

\definecolor{cvprblue}{rgb}{0.21,0.49,0.74}
\usepackage[pagebackref,breaklinks,colorlinks,allcolors=cvprblue]{hyperref}

\def\confName{CVPR}
\def\confYear{2026}

\title{Ego-Pi: VLA Fine-Tuning for Ego-Centric Human and Robot Data}

\author{
\kern-7mm Ji Woong Kim$^{1*}$, Ke Wang$^{1*}$, Zipeng Fu$^{1}$, Sirui Chen$^{1}$, Cong Zhao$^{2}$, Jeff Lai$^{2}$, Chelsea Finn$^{1}$\\
\kern-7mm $^{1}$Stanford University, $^{2}$Meta\\
\textcolor{blue}{\href{https://egopipaper.github.io/}{https://egopipaper.github.io/}}\\
}

\usepackage[flushmargin]{footmisc}

\begin{document}

\twocolumn[{
\renewcommand\twocolumn[1][]{#1}
\maketitle
\begin{center}
    \vspace{-0.26in}
    \centerline{
    \includegraphics[width=\linewidth]{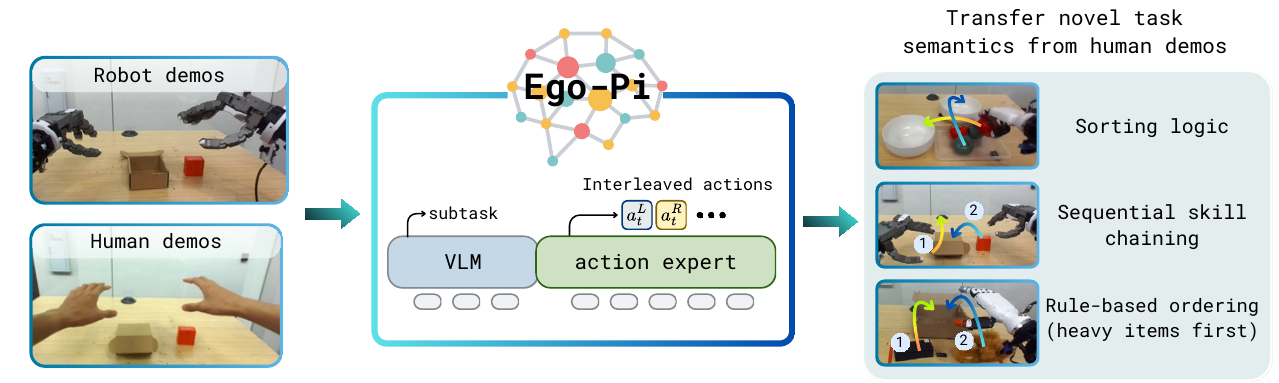}
     }
    \vspace{-0.08in}
   \captionof{figure}{We present Ego-Pi, a framework for cross-embodiment learning from human and humanoid data that adapts gripper-based VLAs for dexterous humanoid control. We investigate how human data can teach robots high-level task semantics, including sorting logic, skill composition, and rule-based ordering.
   }\label{fig:teaser}
   \vspace{-0.02in}
\end{center}
}]

\newcommand\blfootnote[1]{%
  \begingroup
  \renewcommand\thefootnote{}\footnote{#1}%
  \addtocounter{footnote}{-1}%
  \endgroup
}
\begin{abstract}
\blfootnote{%
  \noindent
  \scriptsize
  \begin{tabular}{@{}l}
    $^*$ Denotes co-first authorship. \\
    This is an expanded version of a paper that appeared in CVPR 2026.
  \end{tabular}
}
Robotics faces a fundamental challenge of data scarcity. Unlike language or vision research, there is no internet-scale dataset for robotic manipulation. A promising path forward is to leverage egocentric human data, which can be collected more easily, with greater breadth, and at a larger scale. Towards this end, we investigate key design choices for learning across human and humanoid embodiments equipped with dexterous five-finger hands, using the $\pi_{0.5}$ model as a foundation. Our results show that human data enables robots to learn new task semantics and compose existing skills into novel behaviors without corresponding robot data. 
\end{abstract}    
\section{Introduction}
\label{sec:intro}


Unlike text or vision domains that thrive on internet-scale web data, robot learning is fundamentally bottlenecked by the scarcity of interaction datasets, which traditionally require expensive physical hardware operating in the real world. To overcome this limitation, recent approaches have turned to human data as a scalable alternative. This idea is becoming increasingly viable as robots become more anthropomorphic, equipped with arms and dexterous hands that resemble human morphology. In parallel, the rise of ego-centric devices such as the Apple Vision Pro and Meta Ray-Ban glasses, of which 2 million units sold recently~\cite{EssilorLuxottica2025Q4MetaRayban2mil}, has made large-scale collection of human demonstrations more practical than ever. Together, these developments suggest a new opportunity to scale robot learning by co-training on ego-centric human data.


Towards this goal, we study how to co-train human and robot data effectively using a pre-trained vision–language–action (VLA) model. Specifically, we investigate whether humanoid robots can inherit task semantics and skills that are only present in human data. For instance, we explore whether robots can learn the concept of sorting, skill composition, and rule-based ordering by co-training on such human demonstrations (Fig. \ref{fig:teaser}). This form of cross-embodiment task-semantic transfer has not been explored in depth in prior studies on human–robot co-training, which primarily focus on using human data to improve in-distribution performance or to generalize to novel scenes while performing the same tasks ~\cite{humanpolicy, kareer2024egomimicscalingimitationlearning, lepert2025masquerade, being_h0, egovla}. 


Towards this end, we introduce Ego-Pi (Fig. \ref{fig:teaser}), a framework for adapting pre-trained VLAs to high-dimensional, ego-centric human and humanoid robot data. Ego-Pi addresses a key limitation of existing VLAs: most are developed for gripper-based robots with relatively low-dimensional actions, making them poorly suited for dexterous bimanual control. To overcome this limitation, we use a token interleaving strategy that distributes bimanual actions across two tokens, expanding action capacity without changing the pretrained action head parameters. Since inverse-kinematics-based retargeting is unreliable for high-degree-of-freedom hands, we propose a robot-centric alignment method that directly maps human hand poses into the robot’s joint space.  Finally, we incorporate auxiliary losses, such as subtask generation, to enable stronger task-semantic transfer.

In our experiments, we show that Ego-Pi can reliably learn task semantics only present in human data. For instance, a policy can be steered to sort tomatoes by color, chain existing skills sequentially to perform a boxing task, and learn to perform rule-based packaging, all with success rates of 90\% or higher. We also show that for harder forms of task transfer, such as skill composition, subtask generation as an auxiliary loss is a key ingredient for unlocking good performance. Finally, we find that high-level task-semantic transfer is possible without requiring wrist images for the human data, even while the robot still utilizes wrist observations.

\begin{figure*}[t]
    \centering
    \includegraphics[width=\textwidth]{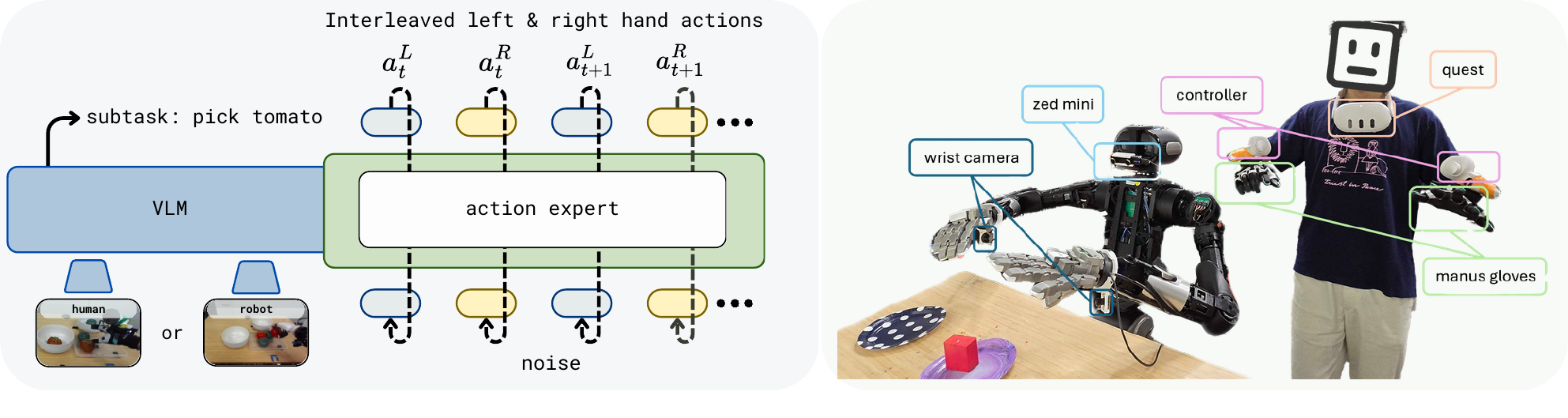}
    \caption{(Left) \textbf{VLA Architecture:} The VLM processes human or robot images to generate bimanual actions interleaved across two tokens per step (left and right hands). This strategy accommodates high-dimensional dexterous control without altering the pretrained action projection layer of the $\pi_{0.5}$ base model. (Right) \textbf{Experimental Setup:} Galaxea R1 Pro robot teleoperated using Manus gloves and Quest controllers for finger and wrist tracking.}
    \label{fig:model_and_setup}
\end{figure*}

\begin{figure}[]
    \centering
    \includegraphics[width=\columnwidth]{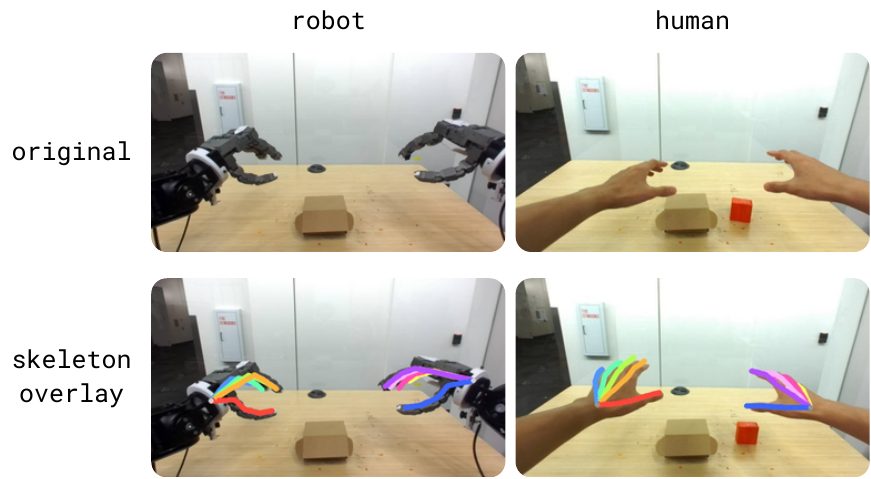}
    \caption{We draw skeleton lines on both the robot and human hands to show how the fingers correspond between them. Each finger is given a unique color that stays the same across both hands. We render the skeleton in a depth-aware, finger-wise manner: fingers that are farther from the camera are drawn behind those that are closer.}
    \label{fig:overlay_comparison}
    \vspace{-1em}
\end{figure}

\section{Related Work}

Robotics research has increasingly explored how human data can help address the limited scale of robot demonstrations. Recent work has used egocentric devices such as Aria glasses~\cite{project_aria} and Apple Vision Pro to extract hand kinematics from human videos. These reconstructed trajectories, together with the accompanying visual observations, provide a rich source of human demonstrations that can be combined with robot data to improve policy performance and generalization.

Several approaches have studied this idea across different robot embodiments. EgoMimic~\cite{kareer2024egomimicscalingimitationlearning} showed that co-training human demonstrations with robot data improves performance, although their robot uses a simple gripper rather than dexterous hands. PH$^{2}$D~\cite{humanpolicy} reported similar benefits on a platform equipped with robot fingers, but their models are trained from scratch and therefore do not benefit from the broader generalization capabilities of foundation models. EgoVLA~\cite{egodex} addressed this limitation by fine-tuning a foundation model on egocentric human data, but they did not evaluate their approach in the real world.

Another line of work seeks to convert human demonstrations directly into robot demonstrations by masking out the human embodiment and overlaying a rendered robot model, as in Masquerade~\cite{lepert2025masquerade} and Mirage~\cite{chen2024miragecrossembodimentzeroshotpolicy}. These pipelines involve many stages and are computationally expensive. Also, the rendered overlays are not occlusion-aware and can obscure the manipulated objects. While these approaches show that human data can improve policy robustness in cluttered scenes, they have not demonstrated the more advanced forms of task-semantic transfer that are the central focus of this paper.

More recently, concurrent work from Physical Intelligence~\cite{human_pi} demonstrated advanced forms of generalization in an egg sorting task, where the robot is trained to sort eggs by color while the human demonstrations specify the sorting rule. However, their method relies on gripper-based robots, and overall task performance remains relatively low. Another concurrent work, EgoScale~\cite{egoscale}, explored teaching robots novel tasks using human data together with a single robot demonstration. In contrast, in our setting we assume access to no robot demonstrations for the target novel task, although that task remains closely related to the robot's existing skills.

\section{Ego-Pi}
\subsection{Preliminaries}

We base our model on the flow-matching vision-language-action (VLA) architecture $\pi_{0.5}$ ~\cite{pi05}, which extends pre-trained vision-language models (VLMs) for robot control. VLAs map multimodal inputs at time $t$, consisting of a language instruction $\ell_t$, images from an ego-centric camera $I_t^{\text{ego}}$, left wrist camera $I_t^{\text{L}}$, and right wrist camera $I_t^{\text{R}}$, together with the proprioceptive state $s_t$, to subtask language output $\hat{\ell}_t$ (optional) and short-horizon continuous actions $a_{t:t+H}$.
Earlier VLA formulations used autoregressive next-token prediction for actions, but $\pi_{0.5}$ adds a flow-matching action head that enables efficient real-time inference.
The model is trained via imitation learning to match expert actions using the following objective:
\begin{equation}
\mathcal{L} = \mathbb{E}_{\tau, \omega} \Big[ \big| \big| \omega - a_{t:t+H} - f_\theta(a_{t:t+H}^{\tau,\omega}, o_t, \ell_t) \big|\big|^2 \Big]    
\label{eq:pi05_loss}
\end{equation}
where $\tau \in [0,1]$ is the flow-time index and $\omega \sim \mathcal{N}(0,I)$ is Gaussian noise.
The interpolated action $a_{t:t+H}^{\tau,\omega} = \tau a_{t:t+H} + (1-\tau)\omega$ represents a partially denoised version of the ground-truth action, and $f_\theta(\cdot)$ predicts the corresponding flow velocity field via the action head.
The observation tuple is defined as $o_t = (I_t^{\text{ego}}, I_t^{\text{L}}, I_t^{\text{R}}, s_t)$, which aggregates all sensory inputs. 


\subsection{Aligning human and robot actions}
\label{sec:align_actions}

To effectively transfer behaviors between human and robot embodiments, it is essential to align their policy action representations. In prior works on hand control ~\cite{egodex, chen2024arcap, humanpolicy} , a common approach to modeling hand actions involves predicting the wrist position and orientation together with the positions of the five fingertips. For robotic systems, this representation must then be translated into desired joint angles to control the robot hand. This conversion is typically performed through inverse kinematics or an optimization-based method ~\cite{chen2024arcap, dexretarget} that regresses joint configurations matching the target fingertip positions.

However, for high-dimensional robot hands, such as the Tesollo hand which is one of the hands used in our setup (20 active joints), these methods often produce self-colliding or unnatural hand poses. To address this issue, we adopt a robot-centric action representation. Specifically, we first convert human hand keypoints, consisting of 20 keypoints across the hand following the MANO convention ~\cite{MANO:SIGGRAPHASIA:2017}, into per-link joint angles, and then map these angles to the robot joint space. This mapping is necessary because the robot hand differs from the human hand in its kinematic structure and proportions. Details of this mapping process are provided in Supplemental Material.

For the robot hand, we directly use its native joint angles as the action representation, since the human hand configuration has already been mapped to the robot joint space. In summary, the policy hand actions are defined as
\begin{equation}
a = \{\, p,\, r,\, q \,\} \in \mathbb{R}^{29},
\end{equation}
where \(p \in \mathbb{R}^3\) denotes the wrist position, 
\(r \in \mathbb{R}^6\) represents the wrist orientation in the 6D rotation format ~\cite{6drot}, 
and \(q \in \mathbb{R}^{20}\) corresponds to the 20 joint angles of the robot hand. For Inspire hands, which is another robot hand we utilize in our experiments, there are only six controllable joints, thus \(q \in \mathbb{R}^{6}\).

\subsection{Adapting gripper-based VLAs for hand control}

Most vision-language-action (VLA) models, including $\pi_{0.5}$, are trained on robots equipped with parallel-jaw grippers, which have relatively low-dimensional action spaces compared to dexterous robot hands. Specifically, the maximum allowed action dimension of $\pi_{0.5}$ is 32, while our dexterous hand consists of 29 dimensions per hand, totaling 58 dimensions for both hands (\cref{sec:align_actions}). The dexterous hand dimensions clearly exceed the 32-dimensional capacity of the original $\pi_{0.5}$ model.

A straightforward approach would be to increase the output dimension of the final projection layer in the $\pi_{0.5}$ action expert. However, in our experiments, this led to higher training losses, likely because it disrupted the original pretrained weights.

To avoid modifying the pretrained weights, we propose an interleaved action sequence formulation for bimanual action prediction, where the left hand and right hand actions are distributed across two action tokens. To illustrate this, let each action at time $t$ be
\begin{equation}
a_t = \{a_t^{L},\, a_t^{R}\},
\end{equation}

where $L$ and $R$ denote the left and right hand respectively. The standard formulation predicts an action sequence of horizon $H$:
\begin{equation}    
a_{t:t+H} = \{\, a_t,\, a_{t+1},\, \ldots,\, a_{t+H} \,\}.
\end{equation}

However, as mentioned above, this approach cannot work because our hand dimensions do not fit within the allowed action dimension of the $\pi_{0.5}$ model. With our proposed interleaving approach, the model emits the left and right actions as separate tokens. In other words, the 58 action dimensions of the dexterous hands are distributed across two 32-dimensional action tokens of the $\pi_{0.5}$ model. Under a fixed action horizon, this allows only $H/2$ bimanual timesteps to be represented:
\begin{equation}
a_{t:t+H/2}
= \{\, 
a_t^{L},\, a_t^{R},\,
\ldots,\,
a_{t+H/2}^{L},\, a_{t+H/2}^{R}
\,\}.
\end{equation}

This interleaving approach preserves the pretrained architecture and its parameters. Although it reduces the effective horizon from $H$ to $H/2$, it enables the model to converge well.
\subsection{Visual alignment of human and robot actions}
While human and robot hands both have five fingers, their visual appearances differ substantially. To make their visual structure comparable, we overlay guiding skeleton lines on the images before feeding them into the model, as illustrated in Fig. \ref{fig:overlay_comparison}. Each finger is assigned a unique color, allowing the model to explicitly understand how fingers from one embodiment correspond to those of the other. The guiding lines are drawn in an occlusion-aware manner so that fingers closer to the camera occlude those farther away. This drawing scheme preserves the interpretability of the hand. Without color coding and occlusion-aware plotting, it becomes difficult to discern which lines correspond to which fingers.

\begin{figure*}[t]
    \centering
    \includegraphics[width=\textwidth]{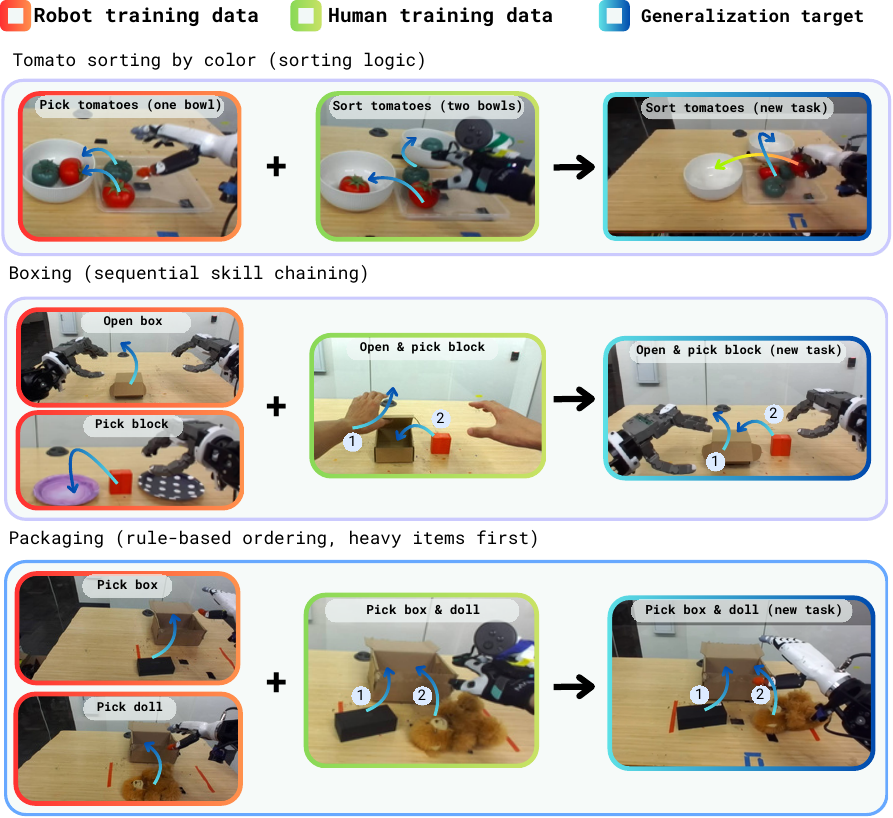}
    \caption{\textbf{Benchmark Tasks.} Evaluation tasks for high-level task-semantic transfer. \textbf{(Top) Tomato Sorting:} Robot data teaches single-bowl placement, while human data introduces multi-bowl sorting. \textbf{(Middle) Skill Composition:} Robot data teaches isolated box-opening and object-placing, while human data teaches sequencing them to place a block inside the box. \textbf{(Bottom) Rule-Based Ordering:} Robot data covers generic item placement, while human data teaches a specific ordering rule during placement (box first, then doll).}
    \label{fig:tasks}
\end{figure*}

\begin{figure*}[t]
    \centering
    \includegraphics[width=\textwidth]{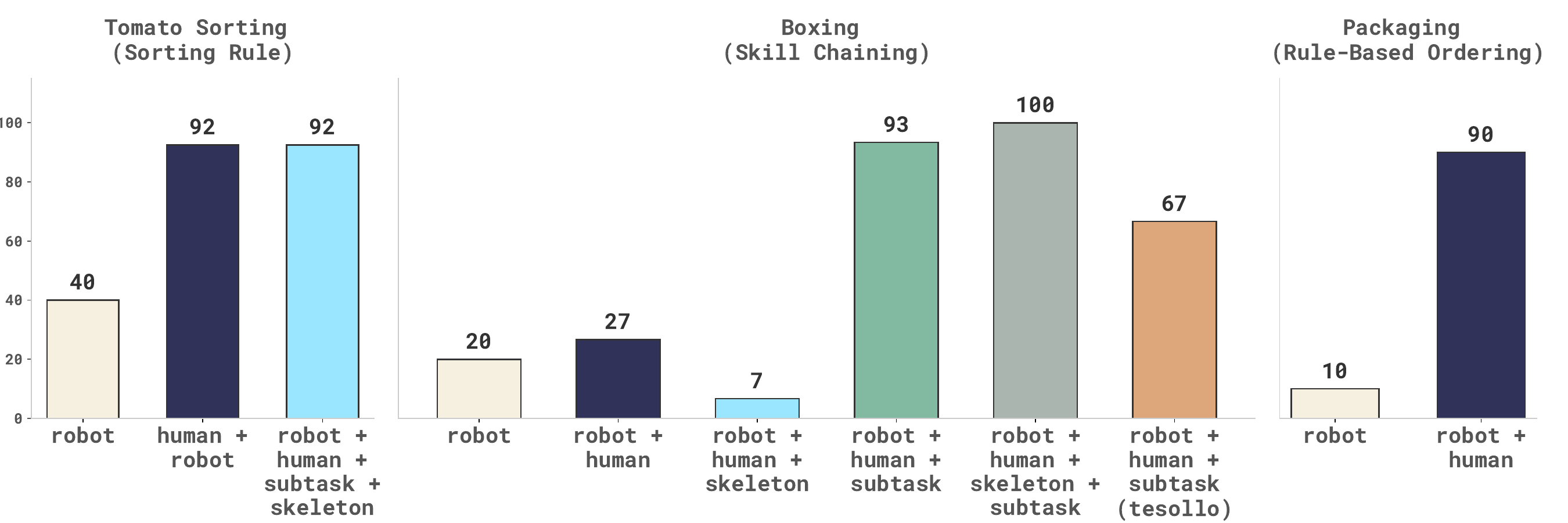}
    \caption{\textbf{Performance across the three tasks.} Co-training with human data is highly effective for the tomato sorting and packaging tasks. For the boxing task, subtask prediction is necessary to achieve high performance. Compared to Tesollo hands, Inspire hands achieve superior performance, likely due to their closer size to human hands. All results are reported using the Inspire hand unless otherwise noted.}
    \label{fig:all_results}
\end{figure*}




\section{Implementation Details}

For the robot setup, we use the Galaxea R1 Pro from Galaxea Dynamics, equipped with a ZED mini mounted on the robot head. Each end-effector is fitted with a Tesollo or Inspire hand, featuring 20 joints or 6 joints per hand, respectively.  Ardurocam wrist cameras are mounted on each wrist, providing a 160° field-of-view. Note that only the robot uses wrist cameras and no wrist cameras are used when collecting human data.

For teleoperation, we utilize Quest controllers mounted on Manus gloves. The Quest controllers track the 6D pose of the operator’s wrists relative to the headset. The robot reproduces the corresponding wrist poses relative to its head using inverse kinematics and PD control. Manus gloves capture the operator’s finger joint angles, which are mapped to the robot’s hand joints to enable dexterous control. Both the Manus gloves and Quest controllers record motion data at 100 Hz. For human data collection, we use a ZED mini camera mounted on a table, and we utilize the same Manus glove and Quest controller setup for human hand tracking. 

During co-training, the ratio between human and robot data is kept at 50\% in a given batch. In practice we found that the policy is not highly sensitive to this ratio. Since human data lacks wrist images, and we wanted to enforce consistent image input across embodiments, the humanoid wrist images were dropped-out 40\% of the time.

Further implementation details are noted in the Supplementary Material section.




\section{Experiments}


We study whether co-training with human data enables the transfer of high-level task semantics absent from robot data. To evaluate this, we design our experiments around the following questions:

\begin{enumerate}
\item How well can simple co-training with human data teach high-level semantics, specifically in tasks that require learning a novel sorting rule, skill composition, and rule-based ordering?

\item How do additional design choices, such as subtask generation or visual augmentation for cross-embodiment alignment, help?

\item How does the choice of the robot hand make a difference in effectively transferring task semantics?

\item How critical is the role of wrist cameras in policy performance?

\end{enumerate}


We design three tasks to systematically evaluate these questions (Fig. \ref{fig:tasks}): tomato sorting by color, boxing, and packaging. The amount of data collected for each task is presented in Table \ref{table}.

\begin{table}[h]
\centering
\small
\caption{Dataset statistics: Number of demos and total duration in minutes for Human (H) and Robot (R) datasets.}
\label{tab:dataset_stats}
\begin{tabular}{lcccc}
\toprule
& \multicolumn{2}{c}{Human (H)} & \multicolumn{2}{c}{Robot (R)} \\
\cmidrule(lr){2-3} \cmidrule(lr){4-5}
Category & \# Demos & Mins & \# Demos & Mins \\
\midrule
Tomato Sort & 89 & 13 & 150 & 60 \\
Boxing & 60 & 5 & 144 & 21 \\
Packaging & 96 & 11 & 185 & 27 \\
\bottomrule
\end{tabular}
\label{table}
\end{table}

\paragraph{Tomato sorting by color} In tomato sorting, we teach the robot the basic skill of placing tomatoes into a single bowl, while human demonstrations introduce the higher-level concept of sorting tomatoes by color into two bowls. By co-training on both sources of data, we evaluate whether the robot can sort tomatoes correctly when presented with two bowls.

\paragraph{Boxing task} In the boxing task, we teach the robot two separate skills: opening a box and picking and placing a block. Human demonstrations then show the direct composition of these skills by opening the box and placing the block inside, thereby boxing. This task evaluates whether co-training enables the robot to compose previously learned skills into a new multi-step behavior.

\paragraph{Packaging} In the packaging task, we teach the robot two separate skills: placing a small box inside a larger box, and placing a bear doll inside the box. Human demonstrations then show the packing rule, which requires the small box to be placed first inside the box and then the bear doll. The idea is that the box is a more rigid and heavier object, therefore it should belong at the bottom. By co-training on both sources of data, we evaluate whether the robot can learn the correct packing order. 

\section{Results}

\paragraph{Q1: How well can simple co-training with human data teach high-level semantics?}

Simple co-training worked well for the tomato sorting and packaging tasks, achieving 92\% and 90\% success rates, respectively. However, it did not perform well on the boxing task. The boxing task requires opening the box before placing the block inside. In practice, the co-trained policy often reached for the block first and attempted to place it on top of the closed box. The policy also sometimes attempted to open the box and pick up the block simultaneously, leading to collisions between the hands. This failure pattern was consistent across both Inspire and Tesollo hands.

In comparison, policies trained solely on robot data performed poorly across all three tasks. For tomato sorting, the success rate dropped to 40\%, and the policy placed tomatoes randomly into the bowls. For the packaging task, the success rate dropped to 10\%; the policy frequently reached for the teddy bear first rather than the smaller box, or awkwardly reached between the two objects instead of clearly targeting a single one. Collisions with the box were also frequently observed. For the boxing task, the success rate dropped to 20\%. Its behavior was similar to the co-trained policy, where the robot reached for the box and the block at the same time, or attempted to place the block inside before the box was even opened.

\paragraph{Q2: How do additional design choices, such as subtask generation or visual augmentation for cross-embodiment alignment, help?} In addition to simple co-training with human data, we explored using subtask prediction and skeleton overlays to improve model performance. Subtask prediction involves labeling the dataset with subtasks and training the VLM component of the VLA to output a subtask string, which serves as an intermediate step before predicting actions via the flow-matching head. The intuition is that by predicting the subtask first, the policy will "think" before acting.

Between the two methods, subtask prediction was the more effective addition, particularly for unlocking high performance in the boxing task. Specifically, incorporating subtask prediction allowed the policy to achieve a 93\% success rate, compared to the 27\% success rate of simple co-training. Utilizing skeleton overlays, however, had little impact on improving policy performance. When we applied both subtask prediction and skeleton overlays to the tomato sorting task, we observed no quantitative or qualitative improvements in performance compared to simple co-training.

One explanation for why the boxing task required subtask prediction to achieve a high success rate is that it requires greater cognitive understanding of the scene. For the tomato sorting and packaging tasks, the policy can simply focus on picking and placing objects into the correct locations. However, the boxing task requires a specific prerequisite condition to be met (i.e., the box being open) before an object is placed inside, making it a more complex scenario. Furthermore, the boxing task is the only bimanual task of the three, requiring careful coordination of both arms that is perhaps more effectively managed through subtask prediction.

\paragraph{Q3: How does the choice of the robot hand make
a difference on effectively transferring task semantics?} We compared the Tesollo and Inspire hands in the boxing and packaging tasks. In the packaging task, the Tesollo and Inspire hands performed at a similar level based on qualitative evaluations. However, when applied to the boxing task, the Tesollo hand struggled to achieve high performance. Similar to the results with the Inspire hand, simple co-training did not work on the Tesollo hand. Even when subtask predictions were applied, the Tesollo hand achieved a 67\% task success rate. Subtask prediction worked well in that the Tesollo hand would always attempt to open the box first instead of reaching for the block, which was a clear improvement. However, it struggled to transition smoothly to the next subtask of placing the block inside the box. Only when the left hand was positioned near a specific location did the next subtask trigger, suggesting that the Tesollo hand policy might be unusually sensitive to the state input. Considering that the Tesollo hand is much larger than a human hand and the Inspire hand is closer to the human form factor, the visual differences between the two may also play a role in downstream performance.

\paragraph{Q4: How critical is the role of wrist cameras in policy performance?} Recall that the human embodiment is trained without wrist cameras and to possibly cope with this difference, the robot wrist cameras are dropped out 40\% of the time during training. Considering that human data only contains third-person images, it may be possible that the policy may pay less attention to the wrist images. At test time, however, when wrist cameras are dropped-out, the policy performance noticeably degraded, indicating that the robot policy is indeed relying on wrist cameras for its performance. For instance, during tomato sorting, if wrist cameras are not provided, the robot struggles to land a stable grasp on the tomato. 

\section{Conclusion and Limitations}

In this work, we demonstrated that teaching high-level task semantics through human data is feasible. We introduced a practical recipe to achieve this, including action interleaving to preserve pre-trained weights, action alignment between human and robot hands, and subtask prediction for optimal results. Together, these components enable the acquisition of novel skills and behaviors that exist only in human data, specifically allowing a humanoid robot to learn novel sorting logic, skill composition, and rule-based ordering with a success rate of 90\% or higher.

Our work has a few limitations. We explored relatively short-horizon, simple pick-and-place-oriented tasks in a fixed-camera scenario. Future work should evaluate the proposed ideas in more challenging scenarios, specifically long-horizon, dexterous manipulation tasks that require mobile manipulation. Additionally, human data should ideally teach robots low-level skills beyond high-level task semantics. In this work, we assumed that the robot was already capable of the necessary low-level skills, using human data primarily as a way to stitch these existing behaviors together in novel ways. Unlocking direct low-level skill transfer remains an open challenge that should be investigated further.

{
    \small
    \bibliographystyle{ieeenat_fullname}
    \bibliography{main}
}

\clearpage
\setcounter{page}{1}
\maketitlesupplementary

\section{Joint Mapping between Human and Robot Hands}

\begin{figure}
    \centering
    \includegraphics[width=\columnwidth]{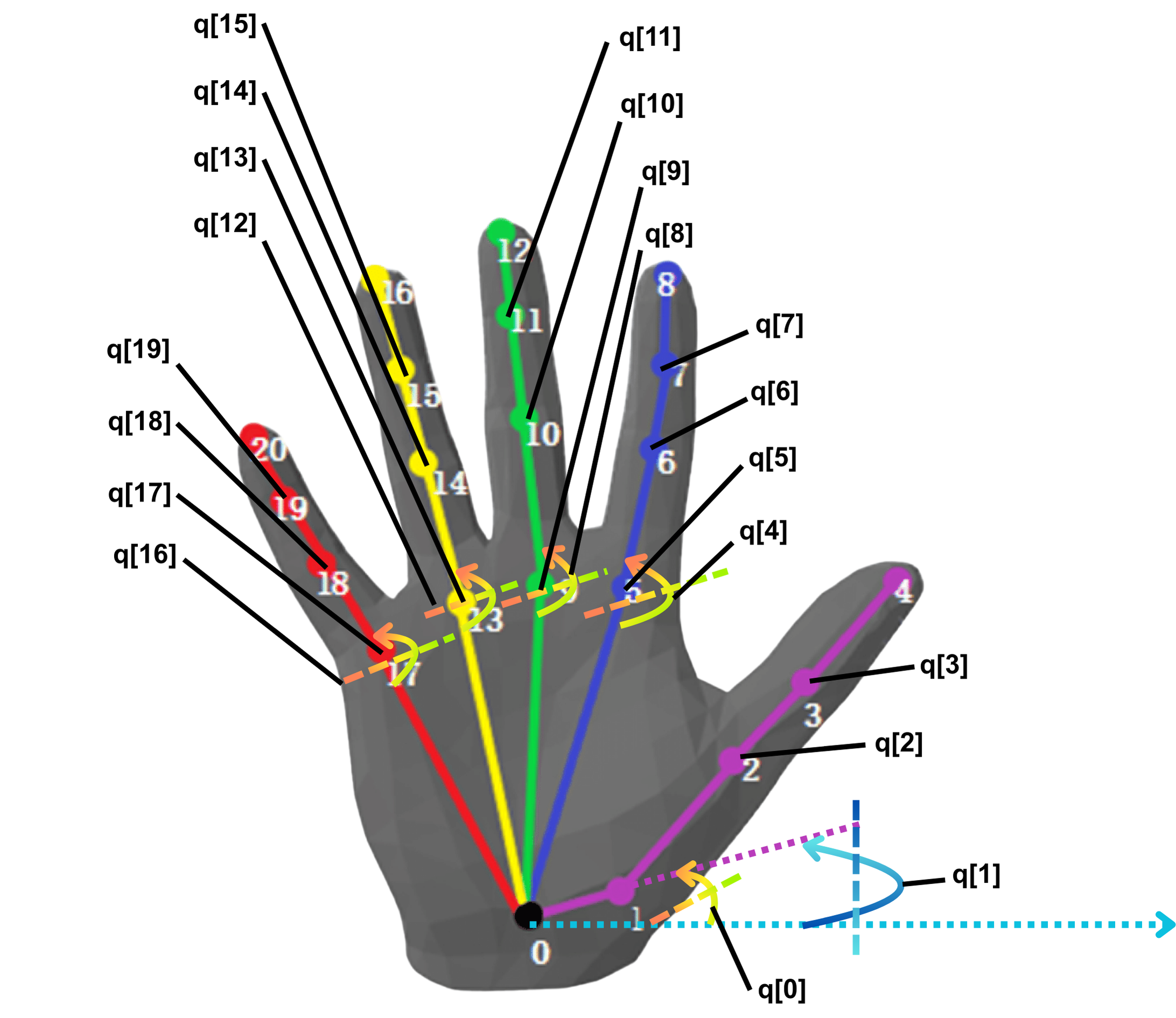}
    \caption{Hand labelled with joint angle locations}
    \label{fig:hand}
\end{figure}

Let $q \in \mathbb{R}^{20}$ be the human hand joint angles measured by the Manus glove (Fig. \ref{fig:hand}), 
and let $q_{\mathrm{robot}} \in \mathbb{R}^{20}$ be the corresponding robot joint angles (e.g., Tesollo hand).

We map each human joint angle to its robot counterpart using a per-joint offset $\delta_i$ and a scaling factor $f_i$:
\begin{equation}
    q_{\mathrm{robot},i}
    = \big(q_i + \delta_i\big)\, f_i,
    \qquad i \in \{1,\ldots,20\}.
\end{equation}

Here, $\delta_i$ is an additive offset and $f_i$ is a multiplicative scaling factor. 
The specific per-joint values used in our implementation are listed below.

\begin{lstlisting}[basicstyle=\footnotesize\ttfamily, breaklines=true]
deg2rad = (pi/180)

q_robot[0] = (38.5 - q[1]) * 0.7 * deg2rad
q_robot[1] = (q[0] + 36.0) * 1.2 * deg2rad
q_robot[2] = (q[2] + 10.0) * 1.2 * deg2rad
q_robot[3] = (q[3] +  5.0) * 1.2 * deg2rad

q_robot[4]  = q[4]  * 1.0 * deg2rad
q_robot[5]  = q[5]  * 1.1 * deg2rad
q_robot[6]  = q[6]  * 1.1 * deg2rad
q_robot[7]  = q[7]  * 1.0 * deg2rad

q_robot[8]  = q[8]  * 1.0 * deg2rad
q_robot[9]  = q[9]  * 1.0 * deg2rad
q_robot[10] = q[10] * 1.0 * deg2rad
q_robot[11] = q[11] * 1.0 * deg2rad

q_robot[12] = q[12] * 1.0 * deg2rad
q_robot[13] = q[13] * 1.0 * deg2rad
q_robot[14] = q[14] * 1.0 * deg2rad
q_robot[15] = q[15] * 1.0 * deg2rad

if q[17] > 55 and q[18] > 25:
    q_robot[16] = abs(q[16]) * 2.0  * 1.0 * deg2rad
else:
    q_robot[16] = abs(q[16]) / 1.5 * 1.0 * deg2rad

q_robot[17] = q[16] * 1.0 * deg2rad
q_robot[18] = q[17] * 1.0 * deg2rad
q_robot[19] = q[19] * 1.0 * deg2rad
\end{lstlisting}

In order to map human hand joint angles provided by HaMeR to robot hand joints, we perform a similar projection with different offsets and scaling factors, as shown below:

\begin{lstlisting}[basicstyle=\footnotesize\ttfamily, breaklines=true]

q_robot[0]  = q[0]
q_robot[1]  = q[1]
q_robot[2]  = q[2]
q_robot[3]  = q[3]

q_robot[4]  = (q[4] - 0.12) * 1.0
q_robot[5]  = q[5]
q_robot[6]  = (q[6]  + 0.18) * 1.1
q_robot[7]  = (q[7]  + 0.18) * 1.1

q_robot[8]  = q[8]
q_robot[9]  = q[9]
q_robot[10] = (q[10] + 0.18) * 1.1
q_robot[11] = (q[11] + 0.18) * 1.1

q_robot[12] = (q[12] + 0.09) * 1.0
q_robot[13] = q[13]
q_robot[14] = (q[14]) * 0.9
q_robot[15] = q[15]

q_robot[16] = (q[16]) * -1.1
q_robot[17] = (q[17] + 0.24) * 1.0
q_robot[18] = (q[18]) * 1.2
q_robot[19] = (q[19]) * 1.3
\end{lstlisting}

\section{Model Details}

The hyperparameters for fine-tuning the $\pi_{0.5}$ policy are shown in Table \ref{tab:hyperparams}.

\begin{table}[h]
\centering
\caption{Hyperparameters for $\pi_{0.5}$ fine-tuning.}
\label{tab:hyperparams}
\begin{tabular}{ll}
\toprule
Hyperparameter       & Value                 \\
\midrule
Optimizer            & AdamW                 \\
$\beta_{1}$          & 0.9                   \\
$\beta_{2}$          & 0.95                  \\
Weight Decay         & 0                     \\
Gradient Clip Norm   & 1.0                   \\
LR Schedule          & Cosine                \\
Warmup Ratio         & 0.001                 \\
Batch Size           & 128                   \\
Training Steps       & 5000 - 10,000                \\
\bottomrule
\end{tabular}
\end{table}
\FloatBarrier

\end{document}